\documentclass[10pt,twocolumn,letterpaper]{article}

\usepackage[pagenumbers]{iccv} %

\let\latexAnd\And
\let\latexAND\AND

\PassOptionsToPackage{numbers, compress}{natbib}

\usepackage[utf8]{inputenc} %
\usepackage[T1]{fontenc}    %

\definecolor{iccvblue}{rgb}{0.21,0.49,0.74}
\usepackage[pagebackref,breaklinks,colorlinks,allcolors=iccvblue]{hyperref}

\usepackage{url}            %
\usepackage{booktabs}       %
\usepackage{amsfonts}       %
\usepackage{nicefrac}       %
\usepackage{xcolor}         %

\usepackage[accsupp]{axessibility}  %
\usepackage{graphicx}
\usepackage{amsmath}

\usepackage{tikz}
\usepackage{tkz-euclide}
\usetikzlibrary{calc}
\usetikzlibrary{math}
\usetikzlibrary{shapes}
\usetikzlibrary{decorations.pathreplacing}

\makeatletter
\renewcommand{\paragraph}{%
    \@startsection{paragraph}{4}%
    {\z@}{0.5ex \@plus .25ex \@minus .2ex}{-1em}%
    {\normalfont\normalsize\bfseries}%
}
\makeatother

\def\paperID{\textsc{12401}} %
\def\confName{ICCV}
\def\confYear{2025}

\title{Modeling Saliency Dataset Bias}

\AtBeginDocument{
    \let\And\latexAnd
    \let\AND\latexAND
}

\author{Matthias Kümmerer\\
\and
Harneet Singh Khanuja\thanks{Now at Georgia Tech}\\
\\
Tübingen AI Center\\
University of Tübingen\\
{\tt\small firstname.lastname@bethgelab.org}
\and
Matthias Bethge\\
}

\begin{document}

\maketitle

\begin{abstract}
Recent advances in image-based saliency prediction are approaching gold standard performance levels on existing benchmarks.
Despite this success, we show that predicting fixations across multiple saliency datasets remains challenging due to dataset bias. We find a significant performance drop (around 40\%) when models trained on one dataset are applied to another. Surprisingly, increasing dataset diversity does not resolve this {\em inter-dataset gap}, with close to 60\% attributed to dataset-specific biases.
To address this remaining {\em generalization gap}, we propose a novel architecture extending a mostly dataset-agnostic encoder-decoder structure with fewer than 20 dataset-specific parameters that govern interpretable mechanisms such as multi-scale structure, center bias, and fixation spread.
Adapting only these parameters to new data accounts for more than 75\% of the generalization gap, with a large fraction of the improvement achieved with as few as 50 samples. Our model sets a new state-of-the-art on all three datasets of the MIT/Tuebingen Saliency Benchmark (MIT300, CAT2000, and COCO-Freeview), even when purely generalizing from unrelated datasets, but with a substantial boost when adapting to the respective training datasets. The model also provides valuable insights into spatial saliency properties, revealing complex multi-scale effects that combine both absolute and relative sizes.

\end{abstract}

\section{Introduction}

Understanding and predicting where we look is valuable for numerous reasons. Scientifically, it provides insights into visual processing in the retina and the brain, memory, emotions and cognitive processes and task driven behaviour. Practically, it enhances applications such as better compression methods, optimized layouts, robotics, and efficient allocation of computational resources.
The field of saliency prediction -- the endeavor of predicting where humans look in images using \textit{saliency models} -- is highly active, with a plethora of different saliency models being developed.

The standard benchmark in the field is the MIT300 dataset~\cite{juddLearningPredictWhere2009} of the MIT/Tuebingen Saliency Benchmark~\cite{kummererMITTuebingenSaliency}.
Recently, the performance on this benchmark has started to plateau~\cite{kummererPredictingVisualFixations2023}, raising questions about whether the field has reached its limits in spatial saliency prediction~\cite{drosteUnifiedImageVideo2020}. 

While there is plenty of room for extensions such as incorporating dynamic eye movements \citep{engbertSpatialStatisticsAttentional2015,adeliModelSuperiorColliculus2017,wlokaActiveFixationControl2018,schwetlickModelingEffectsPerisaccadic2020}, stimulus dynamics \cite{linardosSimpleVsComplex2019,drosteUnifiedImageVideo2020}, or more top-down and task-driven effects \cite{rothkopfTaskContextDetermine2007,chenCOCOSearch18FixationDataset2021,schwetlickDynamicalScanpathModel2023}, spatial saliency remains crucial for many applications.
For concluding that spatial saliency is solved to an extent that is useful for real-world applications, models would need to perform well beyond datasets seen during training. To that end, we conduct a systematic study across five different large saliency datasets and test how well models transfer from one or multiple of these datasets to another. We find a large generalization gap even when training on multiple different saliency datasets.
This indicates that the advantages of training on vast amounts of data -- the key behind the success of recent (foundation) models -- are outweighed by dataset-specific differences.%

To address this issue, we propose a saliency model that is able to account for dataset biases with less than 20 interpretable parameters. Adapting only the bias parameters to new data closes about 75\% of the generalization gap and can be done on as little as 50 samples.
This allows us to substantially improve the state-of-the-art on the MIT300~\cite{juddLearningPredictWhere2009}, CAT2000~\cite{borjiCAT2000LargeScale2015} and COCO-Freeview~\cite{chenCharacterizingTargetAbsentHuman2022} benchmarks in generalization, adaptation and full-training settings.
Furthermore, analyzing the learned dataset specific parameters provides valuable insights into the variability of saliency across datasets.
Our main contributions are as follows:

\begin{itemize}
    \item We identify and quantify large performance penalties when transferring predictions from one or many saliency datasets to an unseen saliency dataset (inter-dataset gap and generalization gap).
    \item We attribute the generalization gap to multiple dataset biases that hinder high saliency prediction performance on unseen datasets: center bias, multiscale distribution, priority scaling and fixation scatter
    \item We propose a new saliency model built on a multiscale backbone architecture which features a simple decoder and incorporates less than 20 interpretable dataset-specific parameters to account for the identified dataset biases.
    \item We demonstrate that adapting only the dataset biases parameters to unseen data successfully closes about 75\% of the generalization gap in a very data-efficient and parameter-efficient way.
    \item We set a new state-of-the-art on the MIT300, CAT2000, COCO-Freeview benchmarks for generalization, adaptation and full training settings, improving AUC performance by at least 1\% compared to previous state-of-the-art on all benchmarks.
    
\end{itemize}

\begin{figure}
    \centering
    \hspace*{-0.3cm}
    \begin{tikzpicture}
        \coordinate (hsep) at (9, 0);
        \coordinate (vsep) at (0, -5);
        \coordinate (labelsep) at (-0.5, -0.0);
        \tikzset{label/.style={font=\sffamily\bfseries}};
        \tikzset{anchor=north west};
        
        \node (IG) at (0, 0) {\includegraphics[width=0.49\textwidth]{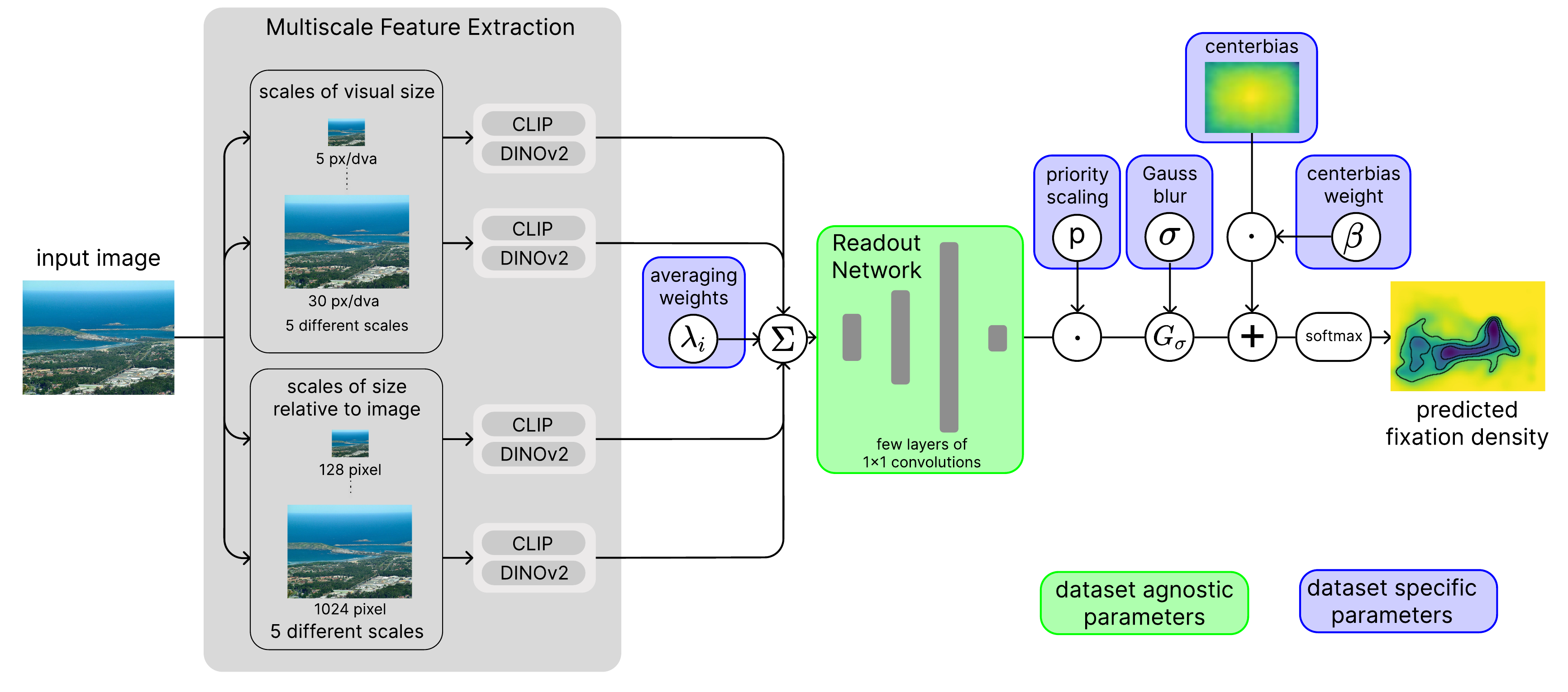}};

    \end{tikzpicture}
    
    \caption{Model Architecture: An input image is rescaled into different resolutions, some defined in total image size in pixels, others in pixels per degree of visual angle. For each image, deep activations from CLIP and DINOv2 encoders are extracted and averaged across scales, from which a priority map is decoded which is then postprocessed with Blur, priority scaling and centerbias. See Appendix Figure \ref{fig:app:architecture} for a larger version.}
    \label{fig:architecture}
\end{figure}

\section{Related Works}

Since Itti and Koch proposed the first image computable saliency model \cite{ittiModelSaliencybasedVisual1998}, a myriad of different saliency models have been proposed.
Early saliency models mostly followed the spirit of the Feature Integration Theory \cite{treismanFeatureintegrationTheoryAttention1980} and proposed that saliency depends on low-level image features such as contrast and edges \cite{torralbaContextualGuidanceEye2006,harelGraphBasedVisualSaliency2007,bruceSaliencyAttentionVisual2009,zhangSUNBayesianFramework2008,kienzleCentersurroundPatternsEmerge2009,richeRARE2012MultiscaleRaritybased2013}.
Later, more and more high-level and semantic image features have been used to predict where people look \cite{juddLearningPredictWhere2009,changFusingGenericObjectness2011} and nowadays in the computer vision community saliency denotes whatever predicts where people look.
The eDN model \cite{vigLargeScaleOptimizationHierarchical2014} introduced deep learning to the field and DeepGaze I \cite{kummererDeepGazeBoosting2015} introduced transfer learning from deep features to improve prediction performance.
Since then, all high-performing saliency models are deep learning based models using some kind of feature transfer from other computer vision tasks \cite{huangSALICONReducingSemantic2015,kruthiventiDeepFixFullyConvolutional2017,kummererUnderstandingLowHighLevel2017,kronerContextualEncoderDecoder2020}. The concept of deep feature transfer has since then been augmented with a variety of techniques to improve prediction performance. 

The SALICON model \cite{huangSALICONReducingSemantic2015} uses a two-scale backbone. Here, we extend and adapt this approach to use a substantially larger number of scales of absolute and relative size, and we don't concatenate the features from multiple scales, but average them.
EML-NET \cite{jiaEMLNETExpandableMultiLayer2020} uses multiple layers from multiple backbones, which is related to our approach of combining deep features from CLIP and DINOv2 backbones. 
The UNISAL model \cite{drosteUnifiedImageVideo2020} trains a saliency model jointly for video and image saliency datasets and uses some domain specific modules to adapt to the different domains. Here we focus only on combining multiple image saliency datasets and analysing the differences within this domain. Furthermore, in UNISAL, most domain specific parameters are not interpretable (batch norm adaptation and linear weights for deep features).
Finally, while both UNISAL and we learn domain or dataset specific parameters, our number of parameters is magnitudes smaller, enabling data-efficient adaptation to new datasets.
SalFBNet \cite{dingSalFBNetLearningPseudosaliency2022} introduced feedback connections from deeper layers of the encoder network to early layers to allow adaptation of low-level features with high-level knowledge. In addition, they pretrained the model on a pseudo saliency dataset obtained by averaging the predictions of top performing saliency models on a large set of images.
DeepGaze IIE \cite{linardosDeepGazeIIECalibrated2021} employed an ensembling strategy (see also \cite{vigLargeScaleOptimizationHierarchical2014}) to average predictions from multiple internal saliency models using different encoder backbones and showed that this results in high performance and good confidence calibration on new datasets. DeepGaze IIE currently represents the state of the art in all datatsets and benchmarks presented in this paper.
Most saliency models use convolutional architectures which keeps the spatial information in a very direct way, more recently also transformer architectures have been introduced \cite{louTranSalNetPerceptuallyRelevant2022,yangPredictingHumanAttention2023,djilaliLearningSaliencyFixations2023}. Here, we combine convolutional and transformer based backbones.
Recently, researchers tried to improve saliency prediction by adding time \cite{aydemirTempSALUncoveringTemporal2024}, Augmentations \cite{aydemirDataAugmentationLatent2025}, explicit global semantic interactions \cite{xieGlobalSemanticguidedNetwork2024} or self-knowledge distillation \cite{termritthikunSalNASEfficientSaliencyprediction2024,termritthikunSemiPKDSemisupervisedPseudoknowledge2025}

The field of eye movement prediction has recognized that different conditions can affect attention behaviour \cite{yarbusEyeMovementsVision1967} and models have been proposed to capture this. However, these works usually included much more different modalities where differences in behavior are more expected (e.g., images/video \cite{drosteUnifiedImageVideo2020} or natural images/webpages and different observer tasks \cite{liUniARUnifiedModel2024}. Opposed to this, \cite{chenWhatDeepSaliency2023} compare different freeviewing image datasets and analyze semantic differences in the fixation distribution, but do not adapt them to new datasets.

The problem of dataset biases has been recognized in other fields of machine learning before \citep{torralbaUnbiasedLookDataset2011,liuDecadesBattleDataset2024} and a variety of adaptation methods tailored more towards large scale datasets have been proposed. Most notable and successful are the test time adaptation methods \citep{schneiderImprovingRobustnessCommon2020,wangTentFullyTestTime2020}. Our approach distinguishes itself from typical test-time adaptation methods in that we build on few mechanisms tailored towards the saliency use case, resulting in few interpretable parameters allowing insights into differences between datasets and highly data-efficient adaptation.

\section{Model}

\paragraph{Overall model architecture}
Our model architecture is visualized in Figure \ref{fig:architecture} %
and can be seen as a variant of the DeepGaze architecture \cite{kummererUnderstandingLowHighLevel2017,linardosDeepGazeIIECalibrated2021}
that has been extended with more modern backbones, a multiscale architecture and where some parameters have been made dataset-dependent.
Firstly, an input image is rescaled into different resolutions. Secondly, for each resolution, deep activations from pretrained backbones are extracted.
Thirdly, the activations from all scales are scaled into a common resolution and combined into a weighted average. Fourthly, the resulting feature maps are decoded into a single spatial priority map with a \textit{readout network} \cite{kummererUnderstandingLowHighLevel2017}, consisting of five layers of 1x1 convolutions. The small number of parameters makes the readout network trainable on datasets with only 1000 images. Lastly, the spatial priority map is multiplied with a priority scaling factor, blurred with a Gaussian and combined with a precomputed center bias log density map which can be down weighted with a weight factor. The result is converted into a probability distribution over pixels using a softmax. %
In the following we detail our extensions to previous DeepGaze architectures. For a precise mathematical formulation and more details on the model architecture see Appendix \ref{app:model}.

\paragraph{Multiscale feature extractor} Our multiscale feature extractor rescales the input image into different resolutions before using the backbones to extract features. Importantly, we use two different sets of scales: The \textit{relative scales} rescale the images to different sizes given in pixels.
The relative scales find patterns that have a certain size relative to the full image. The \textit{absolute scales} on the other hand rescale images to have a fixed resolution in pixel per degree of visual angle (px/dva) and hence depend on how large the image was seen by observers. We call these the absolute scales, because they are sensitive to the absolute visual size of objects independent of the image size. Using both scales together allows to disentangle saliency effects of relative and absolute visual size and build a model that can be applied to new datasets in a meaningful way without having to reason about the right input resolution of an image (opposed to all other compared models).

\paragraph{Pretrained Backbones} For each resolution we extract deep activations from two different pretrained backbones which are then concatenated: CLIP \citep{radfordLearningTransferableVisual2021} is very good at encoding global information about a scene \cite{yuksekgonulWhenWhyVisionLanguage2022,kamathWhatsVisionlanguageModels2023,liuVisualSpatialReasoning2023}, while DINOv2 \citep{oquabDINOv2LearningRobust2023} is designed to encode very precise local information.
Together they should allow to reason about objects within context.

\paragraph{Dataset Biases}
Nearly all parameters of our model are encompassed by the readout network with a total of 26,460 parameters which are trained jointly across datasets.
The remaining parameters are what we call \textit{dataset bias parameters}: they are dataset specific and control interpretable mechanisms.
The dataset bias parameters are: 
the \textit{multiscale averaging weights} which constitute 10 dataset bias parameters that model the relationship between how large the object is both relative to the image but also in terms of how large the object is perceived visually \cite{proulxDoesApparentSize2011};  
 the \textit{priority scaling} models how much more salient, \eg a face is compared to a house,
 and we assume it will vary depending on the experimental conditions, \eg, the engagement of the observers;
the \textit{blur size} specifies the size of the blur kernel (specified in dva) as we assume that datasets might differ in how much fixations are spread out around objects;
the \textit{center bias} constitutes 2-5 dataset bias parameters depending on how exactly the center bias is modeled (see Appendix \ref{app:centerbiases} for more details). It is the spatial prior of the model and encodes a tendency to fixate more towards the center of the image \cite{tatlerCentralFixationBias2007}.
Lastly, the \textit{center bias weight} accounts for the fact that a part of the empirical centerbias from the data, which we use in our models as spatial prior, is explained from image content \cite{einhauserObjectsPredictFixations2008} and hence already predicted by the readout network. Hence, we allow our models to downweight the center bias and make this weight dataset dependent as it depends on the image selection in the datasets.
In total, this amounts to fewer than 20 dataset bias parameters with interpretable values.
We refer to model variants where the bias parameters are identical across datasets as \textit{naive}, as opposed to the full, \textit{bias-aware} model.

\paragraph{Generalization and Adaptation}
Since our model has dataset-specific parameters, applying it to a new dataset requires specifying them. In the generalization setting, we use the average of the dataset bias parameters across all training datasets (including the center bias, where we average the fixation distributions). In the adaptation setting, we finetune the bias parameters on the new dataset.

\section{Experiment Setup}

\paragraph{Datasets}
We use five different datasets in our study: MIT1003 \cite{juddLearningPredictWhere2009}, CAT2000 \cite{borjiCAT2000LargeScale2015}, COCO Freeview \cite{chenCharacterizingTargetAbsentHuman2022,yangPredictingHumanAttention2023}, DAEMONS \cite{schwetlickPotsdamDataSet2024} and FIGRIM \cite{bylinskiiIntrinsicExtrinsicEffects2015}.
Where included, we discard the initial forced fixation.
Unlike all other datasets, FIGRIM technically is not freeviewing data since subjects are doing an image reidentification task where they have to identify repeated image presentations. However, in free-viewing experiments, subjects are commonly told that they will later have to re-identify images to keep engagement up. 
Hence, we assume that in FIGRIM, the eye movement data for the first presentation of each image to a subject is essentially freeviewing data. We exclude eye movement data from repeated image presentations.
We noticed that the CAT2000 dataset contains an artifact,
which we filtered out (see Appendix \ref{app:cat2000artifacts}).
For more details on the used datasets including validation splits see Appendix \ref{app:datasets}.

\paragraph{Loss function and training settings} 
Since our model computes a 2d fixation probability distribution, we can compute average log-likelihood for ground truth fixations by taking the log-probability of the pixel for each fixation and averaging values. Log-likelihood has been shown to be a very powerful loss function for saliency models that generalizes very well to all commonly used saliency metrics 
\cite{kummererInformationtheoreticModelComparison2015,kummererSaliencyBenchmarkingMade2018,kummererPredictingVisualFixations2023}.
All our models are first pretrained on the SALICON dataset \cite{jiangSALICONSaliencyContext2015} and then trained in the actual training setting. 
Overall, we have 4 main training setups: (1) we train one model individually on each dataset and evaluate it on all datasets; for each dataset we train (2) one bias-naive model on the four other datasets (leave-one-out setting) and evaluate the target dataset and (3) one bias-aware model which is evaluated with averaged or adapted dataset parameters; (4) lastly, we train the full model on all five datasets. 
For more details on the training see Appendix \ref{app:training}.

\paragraph{Comparison Models} We include two baseline models. The \textit{centerbias model} is a KDE which, for each image, uses the fixation locations from all other images in the dataset. The centerbias quantifies how well fixations can be predicted without taking image content into account.%
The \textit{gold standard model} estimates inter-observer consistency
and is implemented as suggested by \cite{kummererPredictingVisualFixations2023}, as a regularized KDE which is crossevaluated in a leave-one-subject-out-setting.
In some figures, we specify the gold standard as a range with the upper limit being the mixture of all observer's gold standard models.
For more details on the baseline models, see Appendix \ref{app:baselineModels}.
For comparing to previous state-of-the-art we include DeepGaze IIE \cite{linardosDeepGazeIIECalibrated2021}, UNISAL \cite{drosteUnifiedImageVideo2020}, SalFBNet \cite{dingSalFBNetLearningPseudosaliency2022} and EML-Net \cite{jiaEMLNETExpandableMultiLayer2020}. 
Each of the models is applied in the resolution which resulted in the highest performance on each the dataset. %

\paragraph{Metrics} For our internal analyses and evaluations, we use information gain \cite{kummererInformationtheoreticModelComparison2015,kummererPredictingVisualFixations2023}. Information gain (IG) measures difference in log-likelihood between a candidate model and a baseline model: $IG(\hat p | p_\text{baseline}) = \frac1N\sum_{i=0}^N \left(\log \hat p(x_i \mid I_i) - \log p_\text{baseline}(x_i \mid I_i)\right)$, where $x_i$ are the fixation locations and $I_i$ denotes the image they occured on.
Unlike other metrics, information gain is a ratio scale where differences and ratios of thereof are meaningful, which is needed to answer questions like ``how much of the generalization gap has been closed'' \cite{kummererPredictingVisualFixations2023}. We report information gain relative to the center bias model, quantified in bit per fixation.
For comparing to other models, we furthermore evaluate the commonly used AUC \citep{tatlerVisualCorrelatesFixation2005} metric.
For probabilistic models (DeepGaze IIE, UNISAL and our models) we evaluate AUC on the log densities \cite{kummererSaliencyBenchmarkingMade2018}.

\section{Results}

\begin{figure}
    \centering
    \includegraphics{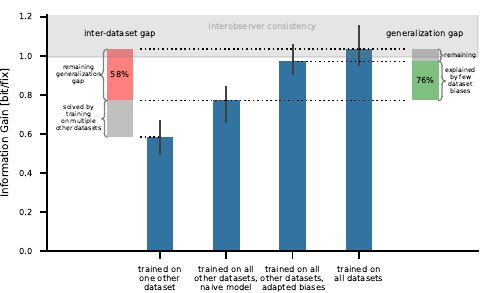}
    \caption{Performing well on unseen datasets is hard due to dataset biases: we show model performances averaged across all five datasets under different training conditions. Generalizing from one dataset to another incurs a substantial performance penalty (``inter-dataset gap''), which largely cannot be fixed by simply training on more other datasets (remaining ``generalization gap''). However, accounting for a few dataset biases parameters and adapting them can mostly close the generalization gap. Performances are mean dataset performances averaged across the five datasets, error bars for paired comparisons are according to \cite{cousineauConfidenceIntervalsWithinsubject2005} and \cite{moreyConfidenceIntervalsNormalized2008}.}
    \label{fig:loo_gaps}
\end{figure}

\begin{table*}
    \caption{Performance of our model and previous state-of-the-art models. Best performance in is indicated in bold, second best is underlined. ``generalization'' refers to training on the respective four other datasets and evaluation with average dataset biases, ``adaptation'' refers to training on the respective four other datasets and evaluation after adapting the dataset bias parameters to the target dataset. Models are sorted by average AUC. See App.~\ref{app:oldModels} for an extended version with more comparison models.}

    \centering
    \resizebox{\textwidth}{!}{%
  \begin{tabular}{lcc@{\hskip 0.2in}cc@{\hskip 0.2in}cc@{\hskip 0.2in}cc@{\hskip 0.2in}cc@{\hskip 0.3in}cc}
    \toprule
    Model & \multicolumn{2}{c}{MIT1003}
          & \multicolumn{2}{c}{CAT2000}
          & \multicolumn{2}{c}{COCO Freeview}
          & \multicolumn{2}{c}{DAEMONS}
          & \multicolumn{2}{c}{FIGRIM}  
          & \multicolumn{2}{c}{average}
          \\
          & IG & AUC
          & IG & AUC
          & IG & AUC
          & IG & AUC
          & IG & AUC
          & IG & AUC \\
    \midrule
      EML-NET & - & 0.842 & - & 0.766 & - & 0.817 & - & 0.766 & - & 0.832 & - & 0.805 \\
SalFBNet & - & 0.883 & - & 0.858 & - & 0.868 & - & 0.774 & - & 0.886 & - & 0.854 \\
UNISAL & 1.006 & 0.887 & 0.099 & 0.865 & 0.712 & 0.873 & 0.712 & 0.809 & 0.771 & 0.892 & 0.660 & 0.865 \\
our model, generalization & 1.172 & 0.902 & 0.249 & 0.878 & 0.889 & 0.886 & 0.538 & 0.800 & 0.883 & 0.905 & 0.746 & 0.874 \\
DeepGaze IIE & 1.113 & 0.894 & 0.315 & 0.878 & 0.846 & 0.881 & 1.006 & 0.822 & 0.877 & 0.899 & 0.831 & 0.875 \\
our model w/o biases, generalization & 1.123 & 0.898 & 0.259 & 0.879 & 0.897 & 0.887 & 0.625 & 0.808 & 0.954 & 0.907 & 0.772 & 0.876 \\
our model, adaptation & \underline{1.217} & \underline{0.904} & 0.469 & 0.887 & 0.965 & 0.890 & 1.149 & 0.840 & 1.059 & 0.911 & 0.972 & 0.886 \\
our model, trained on all & \textbf{1.240} & \textbf{0.905} & \underline{0.522} & \underline{0.891} & \underline{1.031} & \underline{0.895} & \underline{1.258} & \underline{0.848} & \textbf{1.117} & \textbf{0.915} & \underline{1.034} & \underline{0.891} \\
our model, trained per dataset & 1.217 & 0.903 & \textbf{0.535} & \textbf{0.891} & \textbf{1.040} & \textbf{0.895} & \textbf{1.272} & \textbf{0.850} & \underline{1.105} & \underline{0.914} & \textbf{1.034} & \textbf{0.891} \\
\midrule
\textit{Gold Standard (subject-LOO)} & 1.213 & 0.901 & 0.494 & 0.885 & 0.869 & 0.880 & 1.347 & 0.850 & 1.054 & 0.907 & 0.995 & 0.885 \\
\textit{Gold Standard (upper bound)} & 1.829 & 0.945 & 0.873 & 0.920 & 1.511 & 0.935 & 1.722 & 0.899 & 1.642 & 0.947 & 1.515 & 0.929 \\

    \bottomrule
\end{tabular}
}
    \label{tab:main_result}
\end{table*}

\paragraph{Dataset bias parameter adaptation is crucial for improved performance on new datasets}
In \cref{fig:loo_gaps} we show our main results:
Compared to full training on all five datasets, we find that naively generalizing from one dataset to another one results in a substantial performance drop (\textit{inter-dataset gap}) of more than 40\% information gain.
Opposed to what one might expect, this gap cannot be closed by simply training on more diverse data: When training naively on four other datasets, 58\% of the inter-dataset-gap still remains: there is a substantial \textit{generalization gap} to unseen datasets.
By adapting just the dataset-specific parameters of our bias-aware model to the target dataset, we close 76\% of the generalization gap and reach close to full performance.
For all training setups we report averages over mean scores from all five datasets.
In \cref{tab:main_result} we show performances per dataset and also evaluate AUC. In App.~\ref{app:oldModels} we also compare with older saliency models, which on CAT2000 outperform some DNN based models.

\begin{figure}
    \centering
    \includegraphics[scale=0.9]{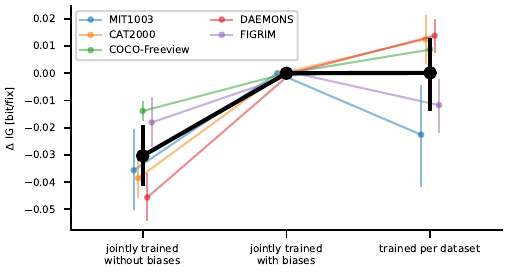}
    \caption{Performing well on multiple datasets is hard without taking dataset biases into account: We compare the performance of our full jointly trained model with dataset biases to the performance of a naively trained model and see that the naive model performs worse.}
    \label{fig:within_domain_biases}
\end{figure}

\paragraph{Adaptation bias parameters is data efficient}
\begin{figure*}
    \centering
    \includegraphics[scale=0.95]{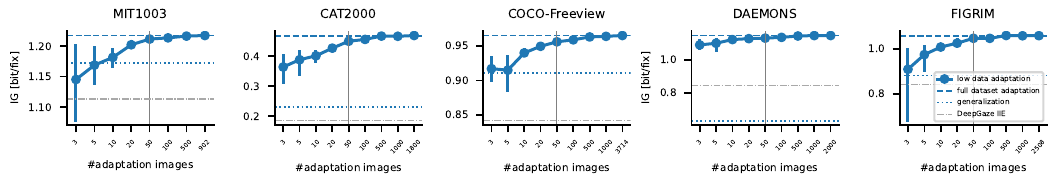}
    \caption{Low data apdatation: adaptation performance depending on the number of images used for adaptation. We outperform generalization with average dataset biases with as little as 5-10 images and reach close to full adaptation performance with around 50 images (vertical lines). Errorbars indicate variance over multiple runs with different random subsets.%
    }
    \label{fig:low_data_adaptation}
\end{figure*}
We test how well we can adapt the dataset bias parameters to unseen datasets when using limited data in our leave-one-dataset-out setting. To this end, we adapt the dataset parameters on small random subsets of the target dataset and evaluate on the full validation split.
Due to the small sample sizes, for the model centerbias, we use a mixture of a KDE, a centered Gaussian and a uniform component (for more details see Appendix \ref{app:centerbiases}).

As shown in \cref{fig:low_data_adaptation} we outperform the generalization case with as little as 5 to 10 images for finetuning. 
With 50 images we already perform close to the full adaptation score. This demonstrates that the dataset bias parameters can successfully be adapted using minimal data, which is an important consideration for real world applications. Since all our dataset bias parameters are interpretable, it may even be possible in some use cases to infer them without any data, yielding performance that surpasses merely averaging the biases of all training datasets.

\paragraph{Dataset bias parameters are critical for achieving strong performance across multiple datasets}

Opposed to the leave-one-out-setting before, in \cref{fig:within_domain_biases}, we compare for each dataset the performance of our jointly trained model with dataset biases with two other training setups: the model is trained jointly on all datasets but without the dataset-specific bias parameters (left) and the model trained separately for each dataset (right). We see that the joint model without dataset-specific bias parameters performs worse on all datasets compared to models trained separately. This finding supports our hypothesis that datasets differ in ways that overshadow the benefits of better learned patterns due to more diverse training data.
Adding dataset specific bias parameters to the jointly trained model compensates for this problem and results in performance comparable to individually trained models. The fact that the joint bias-aware model does not yet outperform individually trained models suggests that additional dataset biases, \eg semantic biases, might be at play which our model does not capture yet. 

Interestingly, the performance drop of the naive jointly trained model is not as large as in the generalization case above. This suggests that the naive model implicitly tries to model some biases by detecting the dataset from the input image and using this information in the saliency decoder, by, \eg computing a center bias from border artifacts in the backbones, which will work only as long as we stay within the training domains.

\paragraph{A new state-of-the-art for free-viewing fixation prediction}

\begin{table*}
    \caption{MIT/Tuebingen Saliency Benchmark: We set a new state-of-the-art on all datasets with generalization, adaptation and full training.}
    
    \centering
    \resizebox{0.8\textwidth}{!}{%
    \begin{tabular}{lcc@{\hskip 0.1in}ccc@{\hskip 0.1in}cccc}
        \toprule
        & \multicolumn{3}{c}{MIT300}
        & \multicolumn{3}{c}{CAT2000}
        & \multicolumn{3}{c}{COCO-Freeview} \\
        
        Model
          & IG & AUC & sAUC
          & IG & AUC & sAUC
          & IG & AUC & sAUC \\
        \midrule
        TempSAL & - & 0.863 & 0.748 & - & 0.844 & 0.638 & - & 0.857 & 0.708 \\
DeepGaze II & 0.951 & 0.876 & 0.784 & 0.084 & 0.864 & 0.650 & 0.664 & 0.870 & 0.740 \\
EML-NET & - & 0.876 & 0.747 & - & 0.831 & 0.585 & - & 0.845 & 0.707 \\
SalFBNet & 0.819 & 0.877 & 0.786 & - & 0.855 & 0.633 & - & 0.872 & 0.710 \\
UNISAL & 0.951 & 0.877 & 0.784 & 0.032 & 0.860 & 0.668 & 0.749 & 0.877 & 0.758 \\
Clueify & - & 0.881 & 0.765 & - & - & - & - & - & - \\
DeepGaze IIE & 1.071 & 0.883 & 0.794 & 0.189 & 0.869 & 0.668 & 0.860 & 0.882 & 0.767 \\
Ours (generalized) & 1.198 & 0.893 & 0.814 & 0.203 & 0.871 & 0.689 & 0.947 & 0.890 & 0.786 \\
Ours (adapted) & \underline{1.236} & \textbf{0.894} & \underline{0.815} & \underline{0.433} & \underline{0.881} & \underline{0.690} & \underline{1.011} & \underline{0.893} & \underline{0.788} \\
Ours (full joint training) & \textbf{1.246} & \textbf{0.894} & \textbf{0.816} & \textbf{0.493} & \textbf{0.885} & \textbf{0.700} & \textbf{1.073} & \textbf{0.897} & \textbf{0.795} \\

        \midrule
        Interobserver consistency
        & 1.3239
        &  	0.8982 
        & -
        & 0.4730
            &  	0.8840
            & 0.6930 
          & 0.8673
            & 0.8829 
            & - 
          \\
        \bottomrule
    \end{tabular}
    }
    \label{tab:benchmarkALL}
\end{table*}

We test our model on the three datasets of the MIT/Tuebingen Saliency Benchmark \cite{kummererMITTuebingenSaliency}: MIT300 (the test set for MIT1003), CAT2000 and COCO-Freeview in three different setups: generalization and adaptation of the bias-aware model trained on the respective four other training datasets, and full joint training.
For all models and datasets,
we set a new state-of-the-art. In particular, with full joint training increasing the main ranking metric AUC by at least 1.1\%--1.5\% in all datasets, with adaptation a close second, emphasizing the power of our modeling approach (\cref{tab:benchmarkALL}, also Appendix \ref{app:benchmark} for all metrics).

\section{Analyses and Ablations}

\paragraph{Importance of different dataset biases:}
\begin{figure}
    \centering
    \includegraphics[scale=0.9]{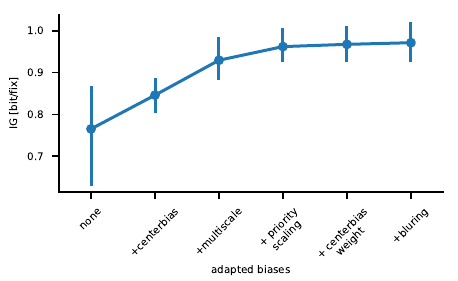}
    \caption{Contribution of different biases to closing the generalization gap. Performances are averages across the five datasets. Error bars \cite{cousineauConfidenceIntervalsWithinsubject2005,moreyConfidenceIntervalsNormalized2008} are quite large because contributions differ across different datasets (Appendix, \cref{fig:loo_bias_per_dataset})}
    \label{fig:ablation_effects}
\end{figure}

In order to understand which of the different dataset biases implemented in our model are most relevant for closing the generalization gap, we conduct an ablation study
in the leave-one-dataset-out setting:
We start evaluating each target-dataset in the generalization setting with averaged bias parameters. We then adapt more and more bias parameters on the target dataset to see how performance increases. Results averaged across all five datasets are shown in \cref{fig:ablation_effects}. We see that centerbias and multiscale weights contribute equally and account for a large part of the performance gain. Priority scaling adds a bit more, the other effects are barely noticable in the dataset average. However, if we compare the performances separate for each dataset (Appendix \cref{fig:loo_bias_per_dataset}), we see that which bias matters how much varies from dataset to dataset. MIT1003 and CAT2000 gain from adapting the centerbias weight, and DAEMONS and FIGRIM gain from adapting the blur size, showing that each bias effect is useful for some datasets.

\paragraph{Case studies:}
\begin{figure*}
    \centering
    \begin{tikzpicture}
        \coordinate (hsep) at (9.5, 0);
        \coordinate (vsep) at (0, -5);
        \coordinate (labelsep) at (-0.5, -0.0);
        \tikzset{label/.style={font=\sffamily\bfseries}};
        \tikzset{anchor=north west};

        \node (IG) at (0, 0) {\includegraphics[scale=0.9]{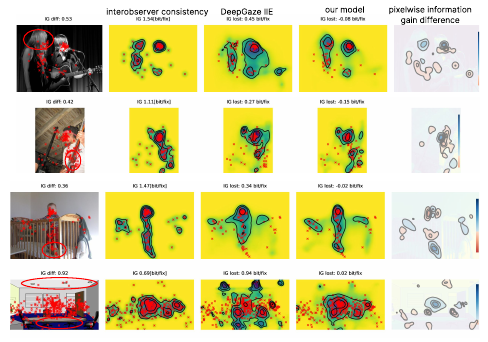}};

        \node (ablation) at ($ (IG.north west) + (hsep) $) {\includegraphics[scale=0.9]{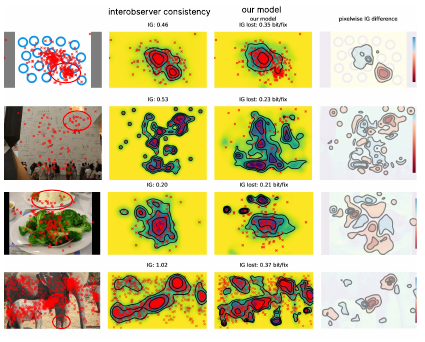}};
        
        \node[label] at ($ (IG.north west) + (labelsep) $) {(a)};
        \node[label] at ($ (ablation.north west) + (labelsep) $) {(b)};
    \end{tikzpicture}
    \caption{Successes and failure cases. (a) Images where our model improves a lot compared to DeepGaze IIE. Areas with notable structure are highlighted: 1.~finds the hidden face, 2.~sees the instrument, 3.~sees the toy animal under foot, 4.~doesn’t get distracted by ceiling lamps and chairs. (b) Images where the model misses patterns: 1.~misses the low-level pattern in lower right (broken circle), 2.~misses the drawing in upper right, 3.~overestimates the salience of dish in background, 4.~misses that people look at an occluded location and overestimates legs. We use \textit{pixelwise information gain difference} \cite{kummererInformationtheoreticModelComparison2015} to visualize where model predictions differ.}
    \label{fig:success_failure}
\end{figure*}
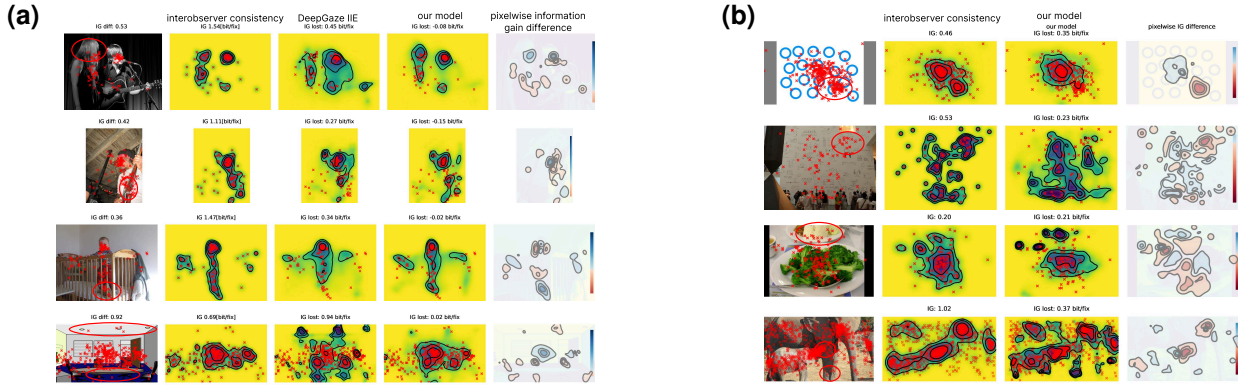
In \cref{fig:success_failure}, we analyse example predictions from cases where our model outperforms the previous state-of-the-art DeepGaze IIE most (success cases), or where it misses most performance compared to inter-observer consistency (failure cases).
We find that our model excels at predicting fixations at hidden faces, objects like instruments and toy animals and gets less distracted by some patterns. This shows that our model not only quantitatively, but also qualitatively improves over DeepGaze IIE.
In terms of failure cases, we find that 
the model underestimates the saliency of low level pattern and abstract drawings, overestimates the salience of background objects and misses fixations on occluded objects. Fixing these issues would both require better high-level understanding of scenes as well as a better understanding of the interplay between high-level and low-level visual features and demonstrates that spatial saliency is still not solved.

\begin{figure*}
    \centering
    \begin{tikzpicture}
        \coordinate (hsep) at (7, 0);
        \coordinate (vsep) at (0, -5);
        \coordinate (labelsep) at (-0.5, -0.0);
        \tikzset{label/.style={font=\sffamily\bfseries}};
        \tikzset{anchor=north west};

        \node (ablationSOTA) at (0, 0) {\includegraphics[scale=0.75]{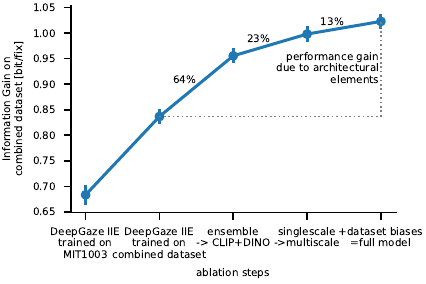}};

        \node (ablationBackbones) at ($ (ablationSOTA.north west) + (6.3, 0) $) {\includegraphics[scale=0.9]{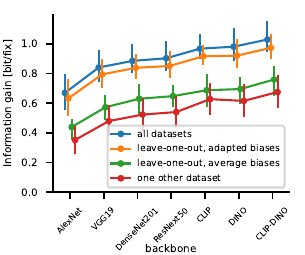}};

        \node (generalizations) at ($ (ablationBackbones.north west) + (5.0,0) $) {\includegraphics[scale=0.9]{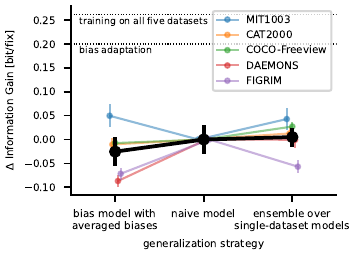}};

        \node[label] at ($ (ablationSOTA.north west) + (0, 0) $) {(a)};
        \node[label] at ($ (ablationBackbones.north west) + (0, 0) $) {(b)};
        \node[label] at ($ (generalizations.north west) + (0, 0) $) {(c)};
        
    \end{tikzpicture}
    \caption{Ablations. (a) Performance gain on the combination of the five dataasets, compared to the previous SOTA model DeepGaze IIE: About half of the performance gain comes from training on the give datasets, the other half is due to architectural changes, comprised of the better backbone, the multiscale architecture and the dataset biases.
    (b) We run our main experiment with many different backbones to confirm that the results about inter-dataset gap, generalization-gap and adaptation hold across backbones.
    (c) Generalizing from a naive model, from a bias-aware model with averaging and via ensembling single-dataset trained models performs roughly on par.
    }
    \label{fig:ablation_sota_backbone_generalization}
\end{figure*}
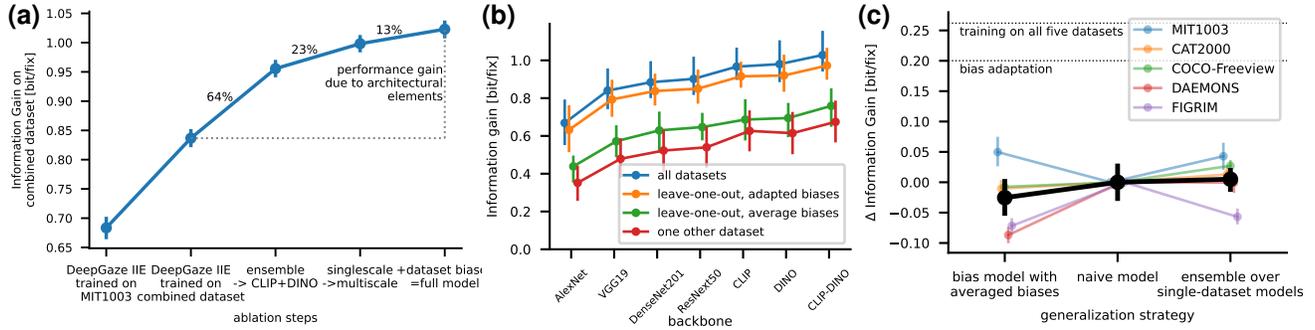

\paragraph{Ablation relative to previous SOTA:}
In \cref{fig:ablation_sota_backbone_generalization}a we compare the performance of different models which step by step transform the previous SOTA model DeepGaze IIE into our new model to quantify the contribution of the different architectural elements. We see that each of the three main architectural changes (replacing the ensemble over multiple backones with a combined CLIP+DINOv2 backbone, adding a multiscale feature extraction and finally making the bias parameters dataset dependent) contributes to the increased performance of our new model.

\paragraph{Dependency on backbone:}
To make sure that our claims about inter-dataset gap, generalization gap and closing thereof with few dataset biases do not depend on our specific backbone (CLIP+DINOv2), we run the same experiments with a variety of different backbones. The backbones are choosen to include the backbones from DeepGaze I \cite{kummererDeepGazeBoosting2015}, DeepGaze II \cite{kummererUnderstandingLowHighLevel2017} and some of the backbones used in DeepGaze IIE \cite{linardosDeepGazeIIECalibrated2021}. The results are shown in \cref{fig:ablation_sota_backbone_generalization}b and confirm that our results hold across backbones.

\paragraph{Different generalization strategies:}
In \cref{fig:ablation_sota_backbone_generalization}c we show that generalization from a naive-model, generalization from the bias-aware model and generalization via ensembling of per-dataset trained models perform roughly on par and far worse than the bias adaptation. Which generalization strategy works best differes from dataset to dataset.

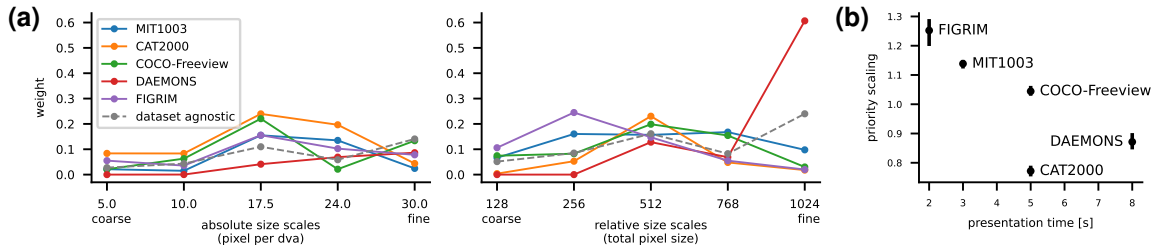
\begin{figure*}
    \centering

    \begin{tikzpicture}
    \coordinate (hsep) at (11, 0);
    \coordinate (vsep) at (0, -5);
    \coordinate (labelsep) at (-0.3, -0.0);
    \tikzset{label/.style={font=\sffamily\bfseries}};
    \tikzset{anchor=north west};

    \node (MultiScale) at (0, 0) {\includegraphics[scale=0.9]{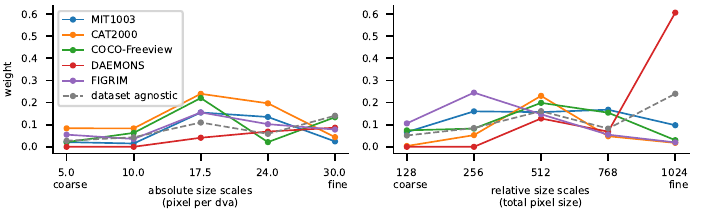}};
    
    \node (PriorityScaling) at ($ (MultiScale.north west) + (hsep) $) {\includegraphics[scale=0.9]{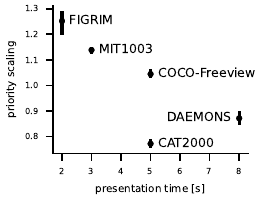}};

    \node[label] at ($ (MultiScale.north west) + (labelsep) $) {(a)};
    \node[label] at ($ (PriorityScaling.north west) + (labelsep) $) {(b)};
    
    \end{tikzpicture}    
    
    \caption{Dataset specific parameters. (a): multiscale weights learned per dataset. For comparison, we also include the weights of the dataset agnostic model. (b) priority scaling learned per datatset as a dependency of presentation time. The reported errors are bootstraped 95\% confidence intervals of the mean across four random seeds.}
    \label{fig:dataset_parameters}
\end{figure*}

\paragraph{Inspection of dataset bias parameters:}
In Figure \ref{fig:dataset_parameters} we show how some of the dataset specific parameters differ across datasets for the model jointly trained across all five datasets. In Figure \ref{fig:dataset_parameters}a we see
 that all datasets require both absolute and relative scales but the specific differ substantially across datasets. DAEMONS requires very high-resolution scales, FIGRIM profits from low-resolution relative scales and most other datasets are roughly in the middle.
In Figure \ref{fig:dataset_parameters}b we show the learned priority scaling parameters across dataset and as a function of presentation time. Again the learned values differ substantially. There is a clear dependency on presentation time, as, e.g., also suggested by \citet{schuttDisentanglingBottomupTopdown2019}, but also variance beyond that (as visible for CAT2000 vs COCO-Freeview).
We also find that the Gaussian blur differs substantially across datasets with DAEMONS requiring much smaller blur size, and that the centerbias weights differ across datasets (Appendix \cref{fig:app:dataset_parameters}).

In Figure \ref{fig:dataset_parameters}a we also show the multi scale weights learned for a dataset agnostic model (dashed line). This shows that saliency follows a complicated multiscale distribution, which requires both information about absolute size (left subplot) as well as relative size (right subplot).%

\paragraph{Further analyses and ablations}
In the Appendix we include additional analyses: We conduct a sensitivity analysis for the bias parameters (Appendix \ref{app:bias_sensitivity_analysis}),
we test generalization and adaptation on the Toronto, Kienzle and SALICON datasets (Appendix \ref{app:newDatasets}),
we explore differences between models trained across datasets or individually (Appendix \ref{app:excels_at_hard_images}),
we assess relevance of different parts of the multiscale architecture in an ablation study (Appendix \ref{app:multiscaleAblation})
and we compare different strategies for modeling the center bias (Appendix \ref{app:centerbiases})

\section{Discussion}
Our results demonstrate that despite the perceived saturation in performance on the MIT300 benchmark, substantial improvements in image-based saliency prediction are still achievable. While training on larger and more diverse datasets is crucial for achieving high performance, we find that dataset-specific biases hinder good performance on unseen datasets and result in a substantial generalization gap. A large part of this gap can be explained with few interpretable mechanisms than can be estimated very data-efficiently on new data, and which are even important to achieve high performance across many training datasets.

Since to a large degree, the bias parameters are likely to depend on the experimental conditions more than on the images themselves, explicitely modeling them is also important for learning good general saliency representations: it avoids that models need to learn shortcuts to essentially make different predictions depending on which dataset an image seems to belong to.
One exception might be the multiscale structure, where we expect also an image dependency, which we hope to explore in the future.

We propose that future saliency research should focus on integrating many available datasets to develop models with robust generalization capabilities. When benchmarking models, it is preferable to evaluate them without retraining on multiple datasets. An interesting approach could be to start reporting aggregated performances over multiple datasets. If models are submitted with their code, benchmarks could evolve towards continual evaluation \cite{prabhuLifelongBenchmarksEfficient2024,ghoshONEBenchTestThem2024}, where new data is regularly added, challenging and potentially reducing the performance of existing models.

This study focused primarily on natural images to remain consistent with typical saliency evaluations. Future work should incorporate a broader range of datasets, including low-level stimuli and other out-of-distribution data. Additionally, the dataset biases considered in this study are limited, and future models could be extended to account for, e.g., varying preferences among different subject cohorts including semantic preferences \cite{haasIndividualDifferencesVisual2019}.

An intriguing outcome of our work is that our new model outperforms estimates of inter-observer consistency on many datasets (\cref{tab:main_result}, \cref{tab:benchmarkALL}). Given that our model still shows some clear failure cases (\cref{fig:success_failure}b), this suggests that the standard methods to estimate per-image inter-observer consistency, usually as KDE \cite{wilmingMeasuresLimitsModels2011,kummererPredictingVisualFixations2023}, may no longer be sufficient. Future efforts should consider new ways to establish a gold standard, such as combining high-performing deep neural networks (DNNs) with models of inter-observer consistency to account for consistent behavior not yet captured by DNNs and guide future model developments.

\paragraph{Acknowledgments} This work was supported by the German Research Foundation (DFG): SFB 1233, Robust Vision: Inference Principles and Neural Mechanisms, project number: 276693517. HSK was supported by the Tübingen AI Center.

\bibliography{literature}

\newpage
\ 
\newpage
\appendix

\section*{Modeling Saliency Dataset Bias - Appendix}

\begin{figure*}
    \centering
    
    \begin{tikzpicture}
        \coordinate (hsep) at (9, 0);
        \coordinate (vsep) at (0, -5);
        \coordinate (labelsep) at (-0.5, -0.0);
        \tikzset{label/.style={font=\sffamily\bfseries}};
        \tikzset{anchor=north west};

        \node (IG) at (0, 0) {\includegraphics[width=0.90\textwidth]{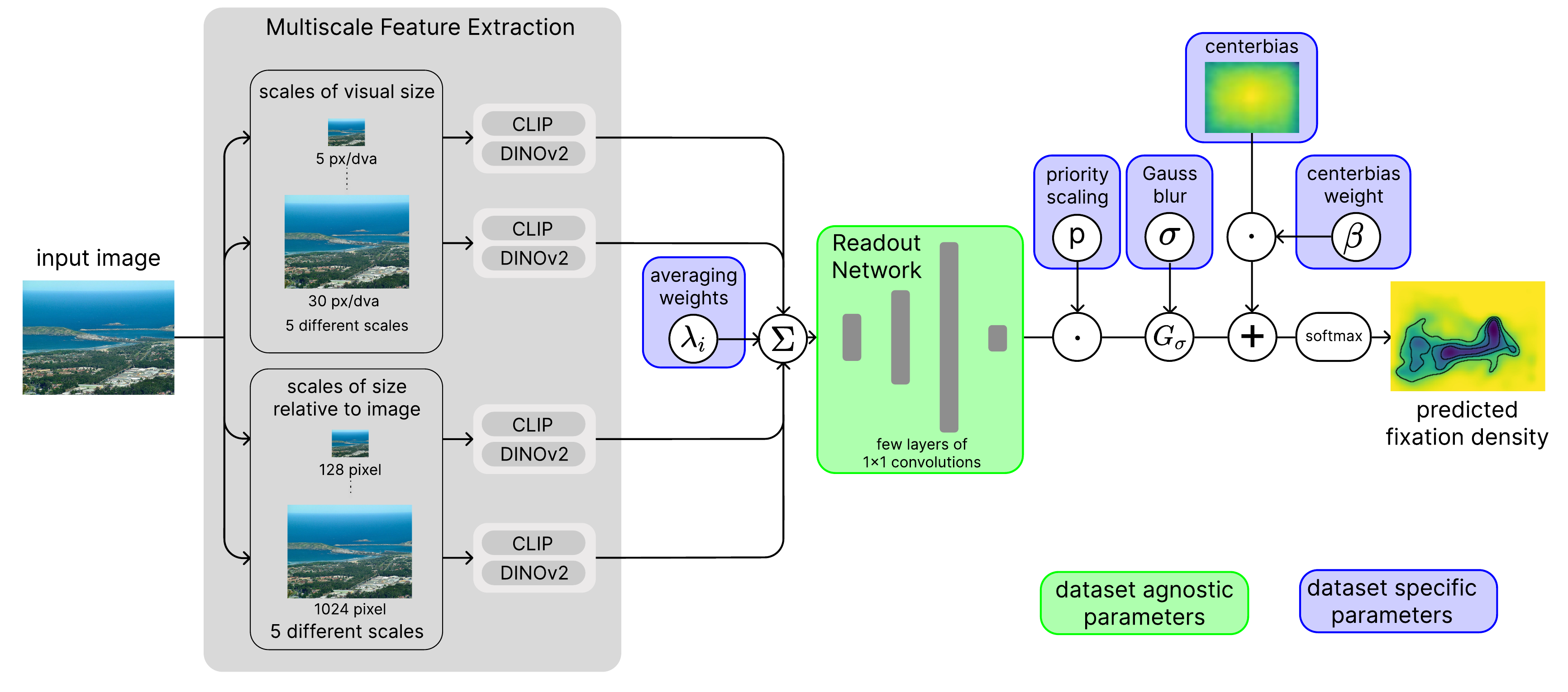}};

    \end{tikzpicture}
    
    \caption{Model Architecture: An input image is rescaled into different resolutions, some defined in total image size in pixels, others in pixels per degree of visual angle. For each image, deep activations from CLIP and DINOv2 encoders are extracted and averaged across scales, from which a priority map is decoded which is then postprocessed with Blur, priority scaling and centerbias}
    \label{fig:app:architecture}
\end{figure*}

\section{Full results on the MIT/Tuebingen Saliency Benchmark}
\label{app:benchmark}

In \cref{tab:benchmarkMIT300}, \ref{tab:benchmarkCAT2000} and \ref{tab:benchmarkCOCOFreeview} we list the full evaluation on the MIT/Tuebingen Saliency Benchmark for MIT300, CAT2000 and COCO-Freeview, including the metrics that we had to skip in the main text due to space reasons.

\begin{table*}
    \caption{MIT300 Benchmark}
    
    \centering
    \begin{tabular}{lccccccc}
        \toprule
        Model & IG & AUC & sAUC & NSS & CC & KLDiv & SIM \\
        \midrule
        SalTR & - & - & 0.7900 & 2.4500 & 0.8000 & 0.3600 & 0.5900 \\
TempSAL & - & 0.8626 & 0.7483 & 2.0092 & 0.7181 & 0.5509 & 0.6202 \\
DeepGaze II & 0.9505 & 0.8759 & 0.7840 & 2.3689 & 0.7851 & 0.4149 & 0.6746 \\
EML-NET & - & 0.8762 & 0.7469 & 2.4876 & 0.7893 & 0.8439 & 0.6756 \\
SalFBNet & 0.8194 & 0.8769 & 0.7858 & 2.4702 & 0.8141 & 0.4151 & 0.6933 \\
UNISAL & 0.9505 & 0.8772 & 0.7840 & 2.3689 & 0.7851 & 0.4149 & 0.6746 \\
GSGNet & - & 0.8780 & 0.7880 & 2.4230 & 0.8110 & 0.4100 & 0.6900 \\
Clueify & - & 0.8811 & 0.7651 & 1.4946 & 0.5750 & 0.8885 & 0.4773 \\
DeepGaze IIE & 1.0715 & 0.8829 & 0.7942 & 2.5265 & 0.8242 & 0.3474 & 0.6993 \\
Ours (generalized) & 1.1975 & 0.8926 & 0.8139 & 2.6697 & 0.8665 & 0.2791 & 0.7311 \\
Ours (adapted) & \underline{1.2355} & \underline{0.8936} & \underline{0.8149} & \underline{2.7229} & \underline{0.8795} & \underline{0.2588} & \underline{0.7478} \\
Ours (full joint training) & \textbf{1.2463} & \textbf{0.8942} & \textbf{0.8159} & \textbf{2.7439} & \textbf{0.8832} & \textbf{0.2540} & \textbf{0.7518} \\

        \midrule
        Interobserver consistency
        & 1.3239
        &  	0.8982 
        & -
        & 2.8481
        & - 
        & - 
        & - \\
        
        \bottomrule
    \end{tabular}
    \label{tab:benchmarkMIT300}
\end{table*}

\begin{table*}
    \caption{CAT2000 Benchmark}
    
    \centering
    \begin{tabular}{lccccccc}
        \toprule
        Model & IG & AUC & sAUC & NSS & CC & KLDiv & SIM \\
        \midrule
        TempSAL & - & 0.8444 & 0.6378 & 1.7037 & 0.6607 & 0.6282 & 0.6173 \\
SalFBNet & - & 0.8549 & 0.6330 & 1.8791 & 0.7028 & 1.2004 & 0.6426 \\
ICF & -0.0229 & 0.8561 & 0.6187 & 1.9588 & 0.7791 & 0.4448 & 0.6697 \\
UNISAL & 0.0321 & 0.8604 & 0.6684 & 1.9359 & 0.7399 & 0.4703 & 0.6633 \\
DeepGaze II & 0.0839 & 0.8640 & 0.6498 & 1.9619 & 0.7950 & 0.3815 & 0.6865 \\
DeepGaze IIE & 0.1893 & 0.8692 & 0.6677 & 2.1122 & 0.8189 & 0.3448 & 0.7060 \\
Ours (generalized) & 0.2031 & 0.8712 & 0.6889 & 2.1460 & 0.8176 & 0.3397 & 0.7200 \\
Ours (adapted) & \underline{0.4333} & \underline{0.8806} & \underline{0.6900} & \underline{2.4591} & \underline{0.8997} & \underline{0.2430} & \underline{0.7726} \\
Ours (full joint training) & \textbf{0.4932} & \textbf{0.8847} & \textbf{0.7002} & \textbf{2.5127} & \textbf{0.9155} & \textbf{0.2098} & \textbf{0.7891} \\

        \midrule
        Interobserver consistency 
        & 0.4730
        &  	0.8840
        & 0.6930 
        & 2.4878
        & -
        & - 
        & - \\
        
        \bottomrule
    \end{tabular}
    \label{tab:benchmarkCAT2000}
\end{table*}

\begin{table*}
    \caption{COCO Freeview Benchmark}
    
    \centering
    \begin{tabular}{lccccccc}
        \toprule
        Model & IG & AUC & sAUC & NSS & CC & KLDiv & SIM \\
        \midrule
        TempSAL & - & 0.8567 & 0.7076 & 1.7508 & 0.6473 & 0.7026 & 0.5626 \\
DeepGaze II & 0.6636 & 0.8699 & 0.7399 & 2.0028 & 0.6909 & 0.5858 & 0.6043 \\
SalFBNet & - & 0.8722 & 0.7099 & 2.0275 & 0.7088 & 0.8623 & 0.6178 \\
UNISAL & 0.7494 & 0.8774 & 0.7585 & 2.0954 & 0.7155 & 0.5515 & 0.6203 \\
DeepGaze IIE & 0.8596 & 0.8825 & 0.7669 & 2.2558 & 0.7563 & 0.4863 & 0.6447 \\
Ours (generalized) & 0.9475 & 0.8896 & 0.7855 & 2.3782 & 0.7907 & 0.4331 & 0.6695 \\
Ours (adapted) & \underline{1.0114} & \underline{0.8932} & \underline{0.7884} & \underline{2.4413} & \underline{0.8048} & \underline{0.4078} & \underline{0.6805} \\
Ours (full joint training) & \textbf{1.0727} & \textbf{0.8968} & \textbf{0.7951} & \textbf{2.5251} & \textbf{0.8258} & \textbf{0.3743} & \textbf{0.6882} \\

        \midrule
        Interobserver consistency 
        & 0.8673
        & 0.8829 
        & - 
        & 2.2837 
        & - 
        & - 
        & - \\
        \bottomrule
    \end{tabular}
    \label{tab:benchmarkCOCOFreeview}
\end{table*}

\section{Comparison with more saliency models}
\label{app:oldModels}

In Table \ref{tab:app:main_result} we extend the results table from the main paper with four additional saliency models, which are not DNN based but take inspiration from neuroscience and psychology: Itti \& Koch \cite{ittiModelSaliencybasedVisual1998} (using the implementation of \cite{harelGraphBasedVisualSaliency2007}), RARE2012 \cite{richeRARE2012MultiscaleRaritybased2013}, GBVS \cite{harelGraphBasedVisualSaliency2007} and CovSal \cite{erdemVisualSaliencyEstimation2013}. From those models, on most datasets, CovSal performs best with the exception GBVS performs substantially better. Interestingly, on CAT2000, CovSal and GBVS perform substantially better than the deep learning model EML-Net, and RARE2012 performs more similarly, but still better than EML-Net.

\begin{table*}
    \caption{Performance of our model and previous state-of-the-art models. Best performance in is indicated in bold, second best is underlined. ``generalization'' refers to training on the respective four other datasets and evaluation with average dataset biases, ``adaptation'' refers to training on the respective four other datasets and evaluation after adapting the dataset bias parameters to the target dataset. Models are sorted by average AUC.}
    
    \centering
    \resizebox{0.9\textwidth}{!}{%
        \begin{tabular}{lcc@{\hskip 0.2in}cc@{\hskip 0.2in}cc@{\hskip 0.2in}cc@{\hskip 0.2in}cc@{\hskip 0.3in}cc}
            \toprule
            Model & \multicolumn{2}{c}{MIT1003}
            & \multicolumn{2}{c}{CAT2000}
            & \multicolumn{2}{c}{COCO Freeview}
            & \multicolumn{2}{c}{DAEMONS}
            & \multicolumn{2}{c}{FIGRIM}  
            & \multicolumn{2}{c}{average}
            \\
            & IG & AUC
            & IG & AUC
            & IG & AUC
            & IG & AUC
            & IG & AUC
            & IG & AUC \\
            \midrule
            Itti \& Koch & - & 0.757 & - & 0.759 & - & 0.702 & - & 0.699 & - & 0.766 & - & 0.736 \\
RARE2012 & - & 0.772 & - & 0.777 & - & 0.771 & - & 0.706 & - & 0.787 & - & 0.762 \\
GBVS & - & 0.803 & - & 0.802 & - & 0.796 & - & 0.710 & - & 0.821 & - & 0.786 \\
CovSal & - & 0.809 & - & 0.847 & - & 0.803 & - & 0.679 & - & 0.835 & - & 0.795 \\
EML-NET & - & 0.842 & - & 0.766 & - & 0.817 & - & 0.766 & - & 0.832 & - & 0.805 \\
SalFBNet & - & 0.883 & - & 0.858 & - & 0.868 & - & 0.774 & - & 0.886 & - & 0.854 \\
UNISAL & 1.006 & 0.887 & 0.099 & 0.865 & 0.712 & 0.873 & 0.712 & 0.809 & 0.771 & 0.892 & 0.660 & 0.865 \\
our model, generalization & 1.172 & 0.902 & 0.249 & 0.878 & 0.889 & 0.886 & 0.538 & 0.800 & 0.883 & 0.905 & 0.746 & 0.874 \\
DeepGaze IIE & 1.113 & 0.894 & 0.315 & 0.878 & 0.846 & 0.881 & 1.006 & 0.822 & 0.877 & 0.899 & 0.831 & 0.875 \\
our model w/o biases, generalization & 1.123 & 0.898 & 0.259 & 0.879 & 0.897 & 0.887 & 0.625 & 0.808 & 0.954 & 0.907 & 0.772 & 0.876 \\
our model, adaptation & \underline{1.217} & \underline{0.904} & 0.469 & 0.887 & 0.965 & 0.890 & 1.149 & 0.840 & 1.059 & 0.911 & 0.972 & 0.886 \\
our model, trained on all & \textbf{1.240} & \textbf{0.905} & \underline{0.522} & \underline{0.891} & \underline{1.031} & \underline{0.895} & \underline{1.258} & \underline{0.848} & \textbf{1.117} & \textbf{0.915} & \underline{1.034} & \underline{0.891} \\
our model, trained per dataset & 1.217 & 0.903 & \textbf{0.535} & \textbf{0.891} & \textbf{1.040} & \textbf{0.895} & \textbf{1.272} & \textbf{0.850} & \underline{1.105} & \underline{0.914} & \textbf{1.034} & \textbf{0.891} \\
\midrule
\textit{Gold Standard (subject-LOO)} & 1.213 & 0.901 & 0.494 & 0.885 & 0.869 & 0.880 & 1.347 & 0.850 & 1.054 & 0.907 & 0.995 & 0.885 \\
\textit{Gold Standard (upper bound)} & 1.829 & 0.945 & 0.873 & 0.920 & 1.511 & 0.935 & 1.722 & 0.899 & 1.642 & 0.947 & 1.515 & 0.929 \\

            \bottomrule
        \end{tabular}
    }
    \label{tab:app:main_result}
\end{table*}

\section{Bias parameter sensitivity analysis:}
\label{app:bias_sensitivity_analysis}
\begin{figure}[b]
    \centering
    \includegraphics{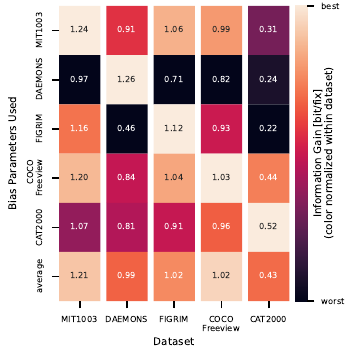}
    \caption{Parameter sensitivity analysis: We evaluate the full model trained jointly across all five dataset on each dataset with the dataset parameters from all datasets and average dataset parameters.}
    \label{fig:parameter_sensitivity_analysis}
\end{figure}
In Figure \ref{fig:parameter_sensitivity_analysis}, we perform a bias parameter sensitivity analysis: for the our foll model trained jointly on all five datasets, we evaluate each dataset in six different settings: once with each set of dataset bias parameters, and once with the averaged dataset bias parameters.  We find that using dataset parameters from a different dataset usually results in a substantial performance drop. This includes the average dataset parameters, however they are usually among the best “wrong” dataset parameters and result in best average performance across datasets. DAEMONS seems to be most different from all other datasets: its dataset parameters result in worst performance on all other datasets. Overall, this analysis shows that the dataset bias parameters control important mechanisms and setting them right can make a large difference in performance.

\section{Generalization and adaptation on new datasets}
\label{app:newDatasets}

\begin{table*}
    \caption{Kienzle Dataset}
    
    \centering
    \begin{tabular}{lccccccc}
        \toprule
        Model & IG & AUC & NSS & CC & KLDiv \\
        \midrule
        EML-NET & - & 0.677 & 0.648 & 0.314 & 1.058 \\
UNISAL & 0.510 & 0.817 & 1.770 & 0.648 & 0.628 \\
DeepGaze IIE & 0.662 & 0.819 & 2.048 & 0.692 & 0.549 \\
our model, average parameters & \underline{1.499} & \underline{0.895} & \underline{2.596} & \textbf{0.879} & \underline{0.284} \\
our model, fine-tuned dataset parameters & \textbf{1.509} & \textbf{0.896} & \textbf{2.603} & \underline{0.879} & \textbf{0.281} \\

        \bottomrule
    \end{tabular}
    \label{tab:kienzle}
\end{table*}

\begin{table*}
    \caption{Toronto Dataset}
    
    \centering
    \begin{tabular}{lccccccc}
        \toprule
        Model & IG & AUC & NSS & CC & KLDiv \\
        \midrule
        EML-NET & - & 0.847 & 2.098 & 0.719 & 2.734 \\
UNISAL & 0.846 & 0.885 & 2.360 & 0.812 & 0.396 \\
DeepGaze IIE & 1.004 & 0.892 & 2.572 & 0.859 & 0.330 \\
our model, average parameters & \underline{1.080} & \underline{0.896} & \underline{2.629} & \textbf{0.879} & \underline{0.284} \\
our model, fine-tuned dataset parameters & \textbf{1.092} & \textbf{0.897} & \textbf{2.640} & \underline{0.879} & \textbf{0.281} \\

        \bottomrule
    \end{tabular}
    \label{tab:app:toronto}
\end{table*}

We test our model on three additional datasets. We use the full joint model trained on all five datasets with dataset bias parameters per dataset. The new datasets are tested both in the generalization setting (averaging the dataset bias parameters including the centerbiases) and adaptation (finetuning the dataset bias parameters on the new dataset).

\paragraph{Kienzle dataset:}
The Kienzle dataset \cite{kienzleCentersurroundPatternsEmerge2009} consists of only 200 images which are random crops of grayscale images of natural scenes, making it an challening testcase.
On this dataset, genrealization already results in substantially improved performance compared to earlier models (8\% in AUC). Adapting the dataset biase parameters to the Kienzle dataset improves performance further (\cref{tab:kienzle}). In Appendix \ref{app:newDatasets} we also test the Toronto dataset \cite{bruceSaliencyAttentionVisual2009}.

\paragraph{Toronto dataset:}
In Table \ref{tab:app:toronto}, we apply our model to the Toronto dataset \cite{bruceSaliencyAttentionVisual2009}. The Toronto dataset consists of 120 images and hence is too small for training full deep learning models which makes it an interesting test case.
On the Toronto dataset, generalization results in improved performance compared to earlier models. Adapting the few dataset parameters to the Toronto dataset improves performance further. Overall, the performance boost, however, is not as large as on the Kienzle dataset. This shows that the Toronto dataset is closer to common saliency datasets and emphasizes the need for new challenging saliency datasets.

\paragraph{SALICON dataset:}
\begin{table}
    \caption{SALICON Dataset}
    \centering
    \begin{tabular}{lc}
        \toprule
        Model & IG \\ %
        \midrule
        Our model, average parameters & -0.03 \\
        Our model, adapted parameters & 0.26 \\
        Our model, trained on SALICON & 0.31 \\
        \bottomrule
    \end{tabular}
    \label{tab:salicon}
\end{table}
We also tested our model on the SALICON dataset. To that end, since our default models all are pretrained on SALICON, we trained a new version of the full model without previous pretraining on SALICON and then again tested generalization and adaptation, comparing to full training on SALICON (\cref{tab:salicon}). We see that the achievable information gain is 0.31 bit/fix. The model trained on our combined dataset (without pretraining on SALICON) and applied with average dataset parameters performs very bad, even slightly worse than the center bias alone. However, adapting the dataset parameters to SALICON results in a performance of 0.26bit/fix, closing 85\% of the generalization gap. This is in line with the results from our leave-one-dataset-out generalization test. In the case of SALICON the dataset parameters seem to account for even more of the generalization gap. This might be due to the extremely different experimental setup of SALICON (e.g., mouse traces instead of eye movements, mechanical turk instead of controlled lab environment).

\begin{figure*}
    \centering
    \begin{tikzpicture}
        \coordinate (hsep) at (5.2, 0);
        \coordinate (vsep) at (0, -5);
        \coordinate (labelsep) at (-0.1, -0.0);
        \tikzset{label/.style={font=\sffamily\bfseries}};
        \tikzset{anchor=north west};

        \node (smallfirst) at (0, 0) {\includegraphics{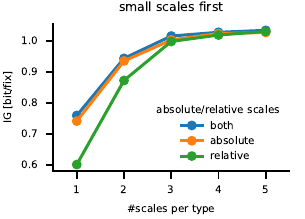}};
        
        \node (largefirst) at ($ (smallfirst.north west) + (hsep) $) {\includegraphics{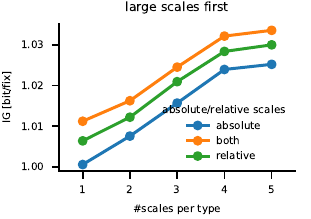}};

        \node (epochtime) at ($ (largefirst.north west) + (hsep) $) {\includegraphics{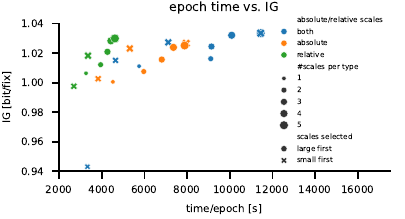}};
        
        \node[label] at ($ (smallfirst.north west) + (labelsep) $) {(a)};
        \node[label] at ($ (largefirst.north west) + (labelsep) $) {(b)};
        \node[label] at ($ (epochtime.north west) + (labelsep) $) {(c)};
        
    \end{tikzpicture}    
    \caption{Multiscale ablation: prediction performance depending on the number of scales in the multiscale backbone and whether the scale weights are dataset agnostic or dataset specific}
    \label{fig:app:multiscale_ablation}
\end{figure*}

\section{Considerations for good centerbias models}
\label{app:centerbiases}

\begin{figure*}
    \centering
    \includegraphics{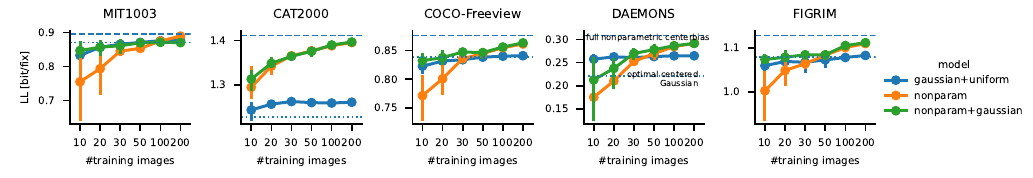}
    \caption{Different centerbias modeling strategies and their performance in low-data settings: we compare three different model classes for centerbias models and how well they perform depending on the number of images used to fit them. Error bars indicate 95\% intervals over multiple runs with different random subsets.}
    \label{fig:app:centerbiases}
\end{figure*}

\begin{figure}
    \centering
    \includegraphics{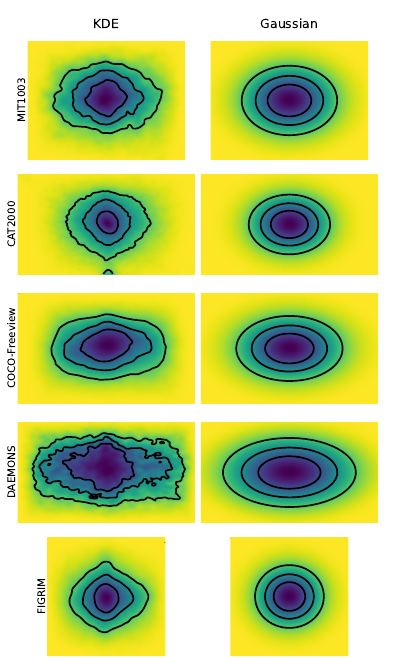}
    \caption{Different centerbias models: We compare a nonparametric centerbias with a centered Gaussian and see that the centered Gaussian often misses a lot of structure present in the average spatial fixation distribution across images.}
    \label{fig:app:centerbiasvisualization}
\end{figure}

On the full datasets, we usually use a centerbias model which is a KDE over all fixations with an additional uniform regularization component. However, in low-data settings, this approach most likely does not average over enough images to result in a good prediction for new images and hence we evaluated different options:
\begin{itemize}
    \item The modeling approach of the full dataset: a KDE with uniform regularization. Bandwidth and regularization weight are selected for maximum likelihood in a leave-one-image-out crossvalidation setting on the training data.
    \item A simple parametric model consisting of a centered Gaussian with a uniform regularization component (this is quite close to what many other models use, \eg \cite{drosteUnifiedImageVideo2020}. Horizontal and vertical variance of the Gaussian as well as the weight of the uniform component are selected to result in maximum likelihood on the training data
    \item A combination of the two previous options: A mixture of a KDE, a centered Gaussian and a uniform component. Horizontal and vertical bandwidth of the Gaussian are computed on the training fixations. The bandwidth of the KDE and the mixture weights are selected for maximum likelihood in a leave-one-image-out crossvalidation setting on the training data.
\end{itemize}

For each of our five datasets and random subsets thereof, we fit the different models and evaluate on the corresponding validation splits. The results are visible in \cref{fig:app:centerbiases}. We see that the first option (``KDE'') sometimes results in bad scores if little data is available. The second option (``Gaussian + uniform'') performs much better in these cases but fails to reach the performance of the nonparametric centerbias with more data. The third option (``KDE+Gaussian+uniform'') combines the advantages of both: reasonable performance already with a few images and good convergence with more data. Interestingly, whether the KDE or Gaussian+uniform performs better for low data is different from dataset to dataset. This is why we select the third option for our low-data adaptation experiments.

These results also serve to demonstrate that modeling the centerbias as a simple Gaussian is not sufficient for many datasets and can result in substantial performance penalties (see also \cref{fig:app:centerbiasvisualization}).

\section{Multiscale ablation}
\label{app:multiscaleAblation}

We evaluated the benefits of our multiscale feature extraction state in an ablation study where we trained the jointly trained model in different settings: we varied whether the model used absolute, relative or both scales. We also varied the numbers of scale per type and whether we added scales starting with the low or high resolutions. We find that the large scales are crucial for performance (Fig.~\ref{fig:app:multiscale_ablation}a, b). We also evaluated computational demand via epoch times and find acceptable performance tradeoffs with fewer but large, preferably relative, scales, resulting in a few percent performance drop (Fig.~\ref{fig:app:multiscale_ablation}c).

\begin{figure*}
    \centering
    \includegraphics{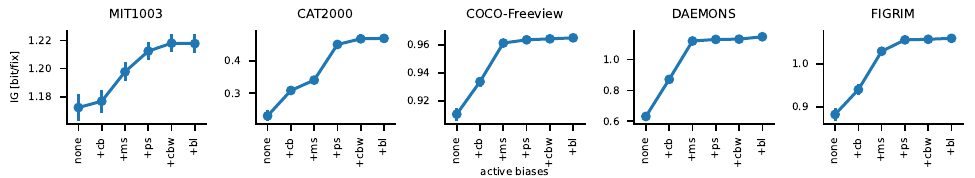}
    \caption{Contribution of different biases to closing the generalization gap, split up target dataset. cb=centerbias, ms=multiscale weights, ps=priority scaling, cbw=centerbias weight, bl=blur size}
    \label{fig:loo_bias_per_dataset}
\end{figure*}

\section{The joint model excels at hard images}
\label{app:excels_at_hard_images}

\begin{figure*}
    \centering
    \includegraphics{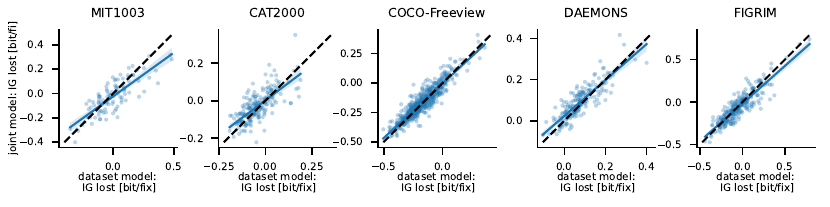}
    \caption{The model trained across dataset tends to perform better at hard images: For each dataset (subplots) we plot prediction error for the individually trained model (x-axis) and the jointly trained model (y-axis) for each image in the validation part of the respective dataset (points). Prediction error is quantified as information gain difference to the gold standard performance. It can be seen that for images where the models trained on only one dataset (right end of the x axis), the joint model tends to perform better than the individually trained model.}
    \label{fig:IGscatterplots}
\end{figure*}

In Figure \ref{fig:IGscatterplots}, we compare the models on a per-image-level. We quantify model performance in terms of the information gain difference to the gold standard, i.e., we measure an prediction error: how much explainable information gain is missed by the the models. This reveals that the joint model profits most from those images where the individual models make the largest prediction error, which means that it performs better on hard images.%

We now analyze model predictions on specific images. From each dataset, we select those images where the jointly trained model outperforms the individually trained models most and visualize the model predictions. In addition, we also visualize the \textit{pixelwise information gain difference} \cite{kummererInformationtheoreticModelComparison2015}: For each pixel, we visualize $p_\text{gold} \left(\log p_\text{joint} - \log p_\text{individual}\right)$. This visualization technique results in highlighting those image areas where predictions differ in a relevant way and makes comparing model predictions more intuitive. For more details on pixelwise information gain, see \cite{kummererInformationtheoreticModelComparison2015}.

\begin{figure*}
    \centering
    \begin{tikzpicture}
        \coordinate (hsep) at (5, 0);
        \coordinate (vsep) at (0, -1.0);
        \coordinate (datasetvsep) at (0, -3.5);
        \coordinate (labelsep) at (-0.5, -0.0);
        \tikzset{label/.style={font=\sffamily\bfseries}};
        \tikzset{anchor=north west};
        
        \coordinate (anchor) at (0, 0);

        \foreach \datasetIndex/\dataset in {0/MIT1003,1/CAT2000,2/COCO-Freeview,3/DAEMONS,4/FIGRIM} {
            \pgfmathsetmacro{\col}{mod(\datasetIndex,3)}
            \pgfmathsetmacro{\row}{div(\datasetIndex,3)}
            
            \coordinate (datasetanchor) at ($ (anchor) + \row*(datasetvsep) + \col*(hsep) $);
            
            \node[anchor=center,label] at ($ (datasetanchor) + (2.5, 0.1) $) {\dataset};

            \foreach \imageNum in {0,1,2} {
                \coordinate (thispos) at ($ (datasetanchor) + \imageNum*(vsep)  $);
                \node at (thispos) {
                    \includegraphics[width=4.3cm]{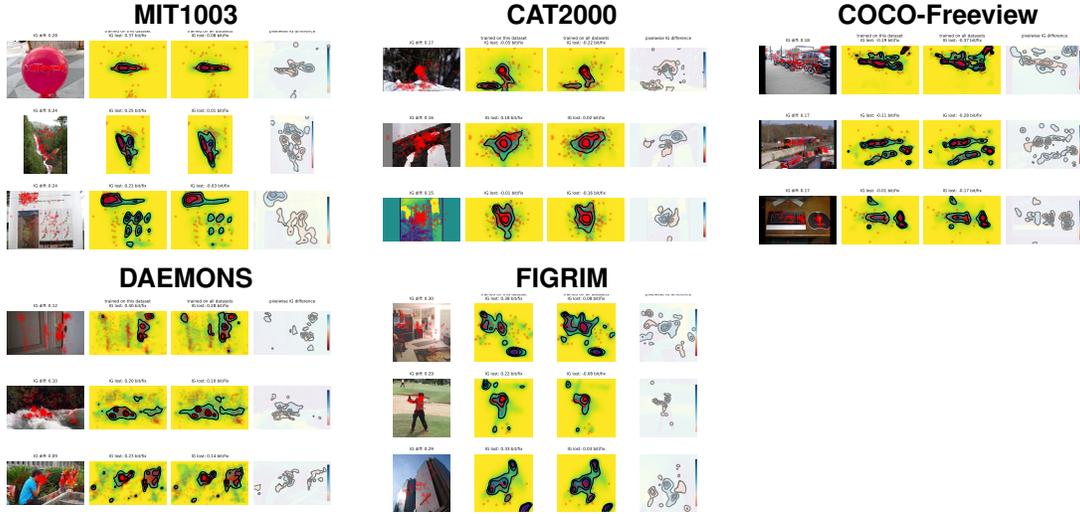}
                };
            }
            
        }

    \end{tikzpicture}
    
    \caption{Example predictions of individually and joint trained models. For each dataset, we show the three images where the difference in performance between the joint model and the individual models is largest. }
    \label{fig:case_study}
\end{figure*}

The resulting images are shown in Figure \ref{fig:case_study}. We see that  the joint model often is better at detecting the exact outline of salient objects (MIT1003, first image), at predicting which one of two salient objects is more important (CAT2000, first image, where more salience is moved to the bird compared to the structure in the foreground; also MIT1003, third image, where the map fragements in the center are downweighted and the peripheral text is upweighted). Also, it appears that the joint model is better at capturing the interplay between local image salience and centerbias, sometimes increasing saliency in the periphery (MIT1003, third image) and sometimes increasing saliency in the center (CAT2000, first image and FIGRIM, first image). Since the model architecture can be seen as computing a posterior from the local image salience as likelihood and the centerbias as prior, this suggests that the joint model manages to extract more evidence from the image features, resulting in overwriting the centerbias more often.

\section{Details about the model architecture}
\label{app:model}

\paragraph{Mathematical model formulation}
Given an image $I$, we denote with $R_{k,j}(I)$ the rescaled images ($k=1, 2$ differentiates between the two scale type ``absolute'' and ``relative'', and $j=1\dots5$ indexes the specific resolutions). We extract deep features $F_{k,j}(I) = F(R_{k,j}(I))$ with our backbone $F$.
Given weights $\lambda_{k,j} \geq 0$, $\sum_{k,j} \lambda_{k,j}=1$ we then compute the averaged deep features $\bar{F} = \sum_{k,j} \lambda_{k,}(R^\prime(F_{k,j}))$, where $R^\prime$ indicates a rescaling operation that rescales all deep features to the same resolution. From $\bar F$, the readout network $RN$ computes a spatial priority map $S=RN(\bar F)$, which is postprocessed with the priority scaling $p$, the Gauss blur size $\sigma$, the center bias distribution $p_\text{cb}(x\mid I)$ and the center bias weight $\beta$ to yield the prediction $\hat p(x\mid I) = \text{softmax}(\mathcal{G}_\sigma(p\cdot S) + \beta \log p_\text{cb}(x\mid I))$.

\paragraph{Multiscale resolutions} We use a total of 10 scales in our multiscale feature extraction. Five scales are resizing the input image to match a certain resolution in terms of pixel per degree of visual angle and use resolutions of 5, 10, 17.5, 24 and 30 px/dva. The other five scales are resizing the input image to match a certain image width or height (whatever is larger) in terms of pixel and uses sizes of 128, 256, 512, 768 and 1024 pixels. Before averaging extracted features across scales, we rescale all of them to 1/8th of the original image resolution to achieve matching sizes. The rescaling operation uses bilinear interpolation.

The scales were chosen to include 17.5 px/dva which is the scale of DeepGaze IIE (MIT1003 has 35px/dva, and DeepGaze IIE downsamples by a factor of 2). From there on we added larger scales until we ran into computational constraints, and smaller scales to the point that we still considered sensible. For the relative scales, the approach was similar: 512 pixel corresponds to the resolution that DeepGaze IIE uses internally on its training dataset, from there on we added smaller and larger scales. In an Ablation (see Appendix \ref{app:multiscaleAblation}), we found that including the larger scales is crucial for improving prediction performance.

\paragraph{CLIP and DINO} We use the implementations and checkpoints from \url{https://github.com/openai/CLIP} and \url{https://github.com/facebookresearch/dinov2}. In the case of CLIP, we use the ResNet50x64 architecture and extract the layer \texttt{layer4.2.conv2}. In the case of DINOv2, we use the ViTB14 architecture and extract the layers \texttt{blocks.6} and \texttt{blocks.10}. In total, this gives us 2560. To regain spatial feature maps from the ViT tokens, we rearange the tokens from the deep layers back into their original layout in the input image. Depending on the image size, we might have differently sized feature maps (for the convolutional CLIP implementation) or different numbers of tokens (for the transformer based DINOv2 implementation). This is not a problem since we don't require the original readout layers and can simply remove them.
The extracted deep layers have been choosen with a random search on MIT1003. Interestingly, we found that for CLIP, the convolutional backbones worked better, while in the case of DINOv2, the ViT based backbones resulted in higher performance. Hence we use the ResNet implementation of CLIP and the ViT implementation of DINOv2.

\paragraph{Readout Network} The readout network consists of five layers of 1x1 convolutions, processing the 2560 feature maps from the multiscale encoder. The five layers of the readout network produce 8, 16, 1, 128, and 1 feature maps respectively. Each layer is prepended by a layer norm and uses softplus as activation function.

\paragraph{Gaussian blur} The output of the readout network is upscaled to 1/2 of the original image resolution before the Gaussian blurring is applied, which is specified in degree of visual angle.

\section{Details about the datasets}
\label{app:datasets}
All datasets except for DAEMONS are accessed via their wrapper in the pysaliency python library \url{https://github.com/matthias-k/pysaliency}. Since DAEMONS is a very new dataset, it's not yet included in pysaliency and we had to write our own pysaliency wrapper. MIT1003, CAT2000 and FIGRIM don't come with official validation splits, here we create our own using \texttt{pysaliency.filter\_datasets.{train,validation}\_fold(stimuli, fixations, crossval\_folds=10, test\_folds=0, val\_folds=1)}. For CAT2000, we furthermore specify \texttt{stratified\_attributes=['category']} to guarantee a uniform distribution of the image categories over splits.

\section{Details about the training}
\label{app:training}

We use the Adam optimizer for optimizing models together with a learning rate schedule consisting of decays of the learning rate by a factor of 10. For each dataset, initial learning rate and points for first and second decay have been selected with a random search. Third and fourth decay always happen after one additional epoch, after the fourth decay training is stopped. The specific learning rate schedules are given in Table \ref{tab:app:learningrates}.

\begin{table*}
  \caption{Learning rate schedule for each dataset}
  \label{tab:app:learningrates}
  \centering
  \begin{tabular}{lll}
    \toprule
    Dataset & initial learning rate & decay epochs \\
    \midrule
    MIT1003 & 0.005623  & 3, 9, 10, 11 \\
    CAT2000 & 0.01 & 6, 9, 10, 11 \\
    COCO Freeview & 0.01 & 12, 15, 16, 17  \\
    DAEMONS & 0.005012 & 12, 15, 16, 17 \\
    FIGRIM & 0.01 & 9, 15, 16, 17\\
    Combined & 0.001585 & 15, 21, 22, 23 \\
    SALICON (pretraining) & 0.01 & 3.75, 7.5, 11.25\\
    \bottomrule
  \end{tabular}
\end{table*}

Pretraining on the SALICON dataset \cite{jiangSALICONSaliencyContext2015}, uses the mouse data from the 2017 SALICON edition. To save compute, for the pretraining we use only one scale with 1024 pixels.

\section{Baseline models}
\label{app:baselineModels}

We include two baseline models to put model performances into perspective: the \textit{centerbias model} is a KDE which, for each image, uses the fixation locations from all other images in the dataset. The centerbias quantifies how well fixations can be predicted without taking image content into account. Bandwidth and a uniform regularization component have been selected for maximum likelihood using leave-one-image-out crossvalidation. For each image in the combined dataset, we use the centerbias prediction from the respective dataset centerbias.

The \textit{gold standard model} estimates inter-observer consistency. As suggested by \cite{kummererPredictingVisualFixations2023}, we use a mixture of a uniform component, the centerbias model and a KDE. The latter uses, for each observer, the fixations from all other observers on the same image. Mixture weights and KDE bandwidth have been chosen for maximum likelihood, where the parameters are fitted for each image individually to make sure that the prediction is as good as possible per image. Unless otherwise indicated, we specify the gold standard performance as 
the leave-on-subject-out crossvalidation performance. For some figures, we specify the gold standard as 
a range ranging from the leave-on-subject-out crossvalidation performance up to the performance when including all image fixations in the KDE but keeping the parameters fitted in the crossvalidation. The first is essentially a lower bound on inter-observer consistency, the latter is an upper bound.

\section{CAT2000 artifacts}
\label{app:cat2000artifacts}

\begin{figure*}
  \centering
  \includegraphics[scale=0.9]{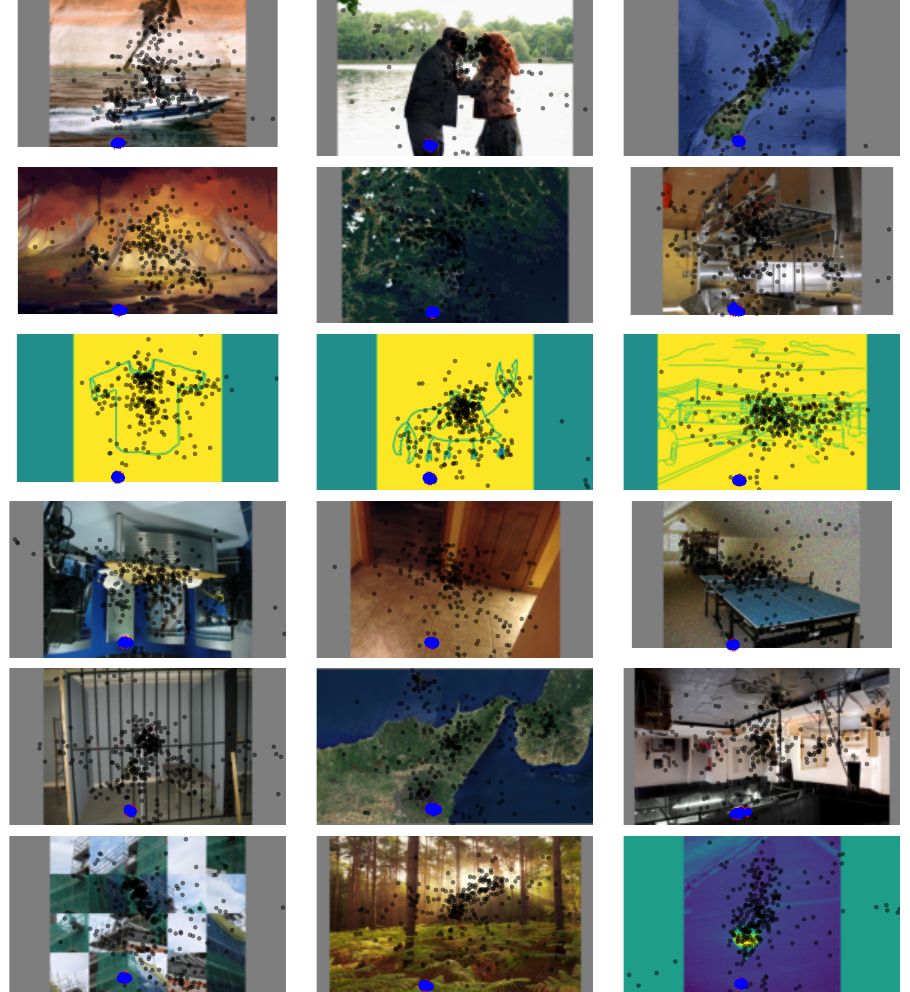}
  \caption{Artifacts in the CAT2000 dataset: some scanpaths have all fixations clustered in always the same image area. This likely indicates eye tracking problems and hence we excluded the scanpaths. The fixtions of the respective scanpath are shown in blue, for comparison all fixations from other subjects are additionaly shown in black.}
  \label{fig:app:cat2000_artifacts}
\end{figure*}

We noticed that the CAT2000 dataset contains an artifact: for some scanpaths, all fixations are all clustered in a small image area far from the image center. All these scanpaths are from the same subject, indicating eye tracking problems with this subject. For this reason, we excluded all these scanpaths from the dataset by removing all scanpaths from subject number 20 with a mean y position of larger than 950 pixels. Extensive visual tests confirmed that this indeed removes those and only those scanpaths (see Figure \ref{fig:app:cat2000_artifacts} for example cases)

\begin{figure*}
    \centering
    
    \begin{tikzpicture}
        \coordinate (hsep) at (6, 0);
        \coordinate (vsep) at (0, -5);
        \coordinate (labelsep) at (-0.5, -0.0);
        \tikzset{label/.style={font=\sffamily\bfseries}};
        \tikzset{anchor=north west};

        \node (MultiScale) at (0, 0) {\includegraphics{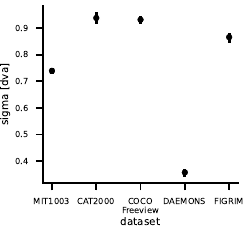}};
        
        \node (PriorityScaling) at ($ (MultiScale.north west) + (hsep) $) {\includegraphics{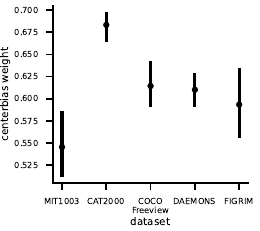}};

        \node[label] at ($ (MultiScale.north west) + (labelsep) $) {(a)};
        \node[label] at ($ (PriorityScaling.north west) + (labelsep) $) {(b)};
        
    \end{tikzpicture}    
    
    \caption{Dataset specific parameters. (a): Bandwidth of Gaussian blur per dataset in degree of visual angle (dva). (b) centerbias weight per dataset. The reported errors are bootstraped 95\% confidence intervals of the mean across four random seeds.}
    \label{fig:app:dataset_parameters}
\end{figure*}

\section{Assets}

Our models where implemented in python using pytorch \citep{paszkePyTorchImperativeStyle2019}. Model evaluations and saliency metrics were using the public pysaliency toolbox (\url{github.com/matthias-k/pysaliency}, MIT license). All datasets except for DAEMONS were used via their pysaliency wrapper. The models were used via their implementations from \url{https://github.com/rdroste/unisal} (Apache 2 license), \url{https://github.com/SenJia/EML-NET-Saliency}, \url{https://github.com/gqding/SalFBNet} and \url{https://github.com/matthias-k/deepgaze}. Also used were scipy \cite{virtanenSciPy10Fundamental2020} and numpy \cite{harrisArrayProgrammingNumPy2020} for computations, pandas \citep{rebackPandasdevPandasPandas2022} for statistics and data handling as well as matplotlib \citep{hunterMatplotlib2DGraphics2007} and seaborn \cite{waskomSeabornStatisticalData2021} for plotting.

\section{Compute Ressources}

All main experiments where conducted on A100s. Model trainings on individual datasets took around 6--24 hours, trainings on the combined dataset around 3-4 days. The learning rate random search was conducted using an earlier model version on 2080Ti GPUs. Around 500 random search iterations were performed taking on average 5 hours each.

\bibliography{literature}

\end{document}